\documentclass[sigconf]{acmart}

\AtBeginDocument{%
  \providecommand\BibTeX{{%
    \normalfont B\kern-0.5em{\scshape i\kern-0.25em b}\kern-0.8em\TeX}}}

\setcopyright{acmcopyright}
\copyrightyear{2018}
\acmYear{2018}
\acmDOI{XXXXXXX.XXXXXXX}

\acmConference[Conference acronym 'XX]{Make sure to enter the correct
  conference title from your rights confirmation emai}{June 03--05,
  2018}{Woodstock, NY}
\acmPrice{15.00}
\acmISBN{978-1-4503-XXXX-X/18/06}

\usepackage[acm]{definition}

\newcommand{\ie}{\emph{i.e.}\xspace} % that is
\newcommand{\eg}{\emph{e.g.\xspace}} % for example

\newcommand{\nop}[1]{}

\newcommand{\method}{\textsc{Weapo}\xspace}

\begin{document}

\title{Binary Classification with Positive Labeling Sources}

%
% The "author" command and its associated commands are used to define
% the authors and their affiliations.
% Of note is the shared affiliation of the first two authors, and the
% "authornote" and "authornotemark" commands
% used to denote shared contribution to the research.
\author[Jieyu Zhang, Yujing Wang, Yaming Yang, Yang Luo, and Alexander Ratner]{Jieyu Zhang$^{1,2}$, Yujing Wang$^1$, Yaming Yang$^1$, Yang Luo$^1$, Alexander Ratner$^{2,3}$}
\affiliation{
	\institution{$^1$Microsoft Research Asia}
	\institution{$^2$The Paul G. Allen School of Computer Science \& Engineering, University of Washington}
	\institution{$^3$Snorkel AI, Inc.}
	\country{}
}

\email{{jieyuz2, ajratner}@cs.washington.edu , {yujwang, yayaming,yangluo}@microsoft.com}

\def\authors{Jieyu Zhang, Yujing Wang, Yaming Yang, Yang Luo, and Alexander Ratner}

\renewcommand{\shortauthors}{Zhang, et al.}

\begin{abstract}
To create a large amount of training labels for machine learning models effectively and efficiently, researchers have turned to Weak Supervision (WS)~\cite{Ratner16}, which uses programmatic labeling sources rather than manual annotation.
Existing works of WS for binary classification typically assume the presence of labeling sources that are able to assign both positive and negative labels to data in roughly balanced proportions.
However, for many tasks of interest where there is a minority positive class, negative examples could be too diverse for developers to generate indicative labeling sources.
Thus, in this work, we study the application of WS on binary classification tasks with positive labeling sources only.
We propose \method, a simple yet competitive WS method for producing training labels without negative labeling sources.
On 10 benchmark datasets, we show \method achieves the highest averaged performance in terms of both the quality of synthesized labels and the performance of the final classifier supervised with these labels. We incorporated the implementation of \method into WRENCH~\cite{zhang2021wrench}, an existing benchmarking platform\footnote{\url{https://github.com/JieyuZ2/wrench/blob/main/wrench/labelmodel/weapo.py}}.
\end{abstract}






\maketitle

\section{Introduction}

Weak Supervision (WS), one recent paradigm~\cite{Ratner16} for overcoming the challenge of low availability of training labels, has achieved remarkable success in various real-world applications~\cite{bach2019snorkel,Fries2019WeaklySC}.
Specifically, in WS, expensive, time-consuming manual annotations are replaced with programmatically-generated labeling sources, called \emph{labeling functions (LFs)}, applied to unlabeled data.
The usable labeling sources include but not limited to heuristics, knowledge bases, and pretrained models- they are typically cheaper than hand-labeling and able to provide potentially noisy labels to unlabeled data at scale. 
Despite the efficacy and efficiency of WS framework, it heavily relies on high-quality labeling sources to achieve satisfactory performance~\cite{zhang2021wrench,zhang2022understanding}.
In some real-world scenarios, in particular ones with a minority "positive" and majority "negative" class, the data of certain label could be highly  diverse, making it difficult for practitioners to create labeling sources.
For example, in bot detection where the user aim to detect bots from normal users, it is non-trivial even for experts to create labeling sources that identify the patterns of normal users, since the user behavior could be arbitrary.
Note that such a motivation indeed drives the long-standing research problem of Positive-Unlabeled learning~\cite{DBLP:journals/corr/abs-1811-04820}.
With regard to the importance of labeling sources and the difficulty of generating them for certain applications, in this work, we study the effectiveness of WS approaches under the setting where the labeling source of certain data class is absent.
Specifically, we focus on binary classification with positive labeling sources only; in other words, the labeling source at-hand could only assign positive or abstain on any given data point.

We propose \method, a simple yet effective method for binary classification with positive labeling sources only. 
In particular, it is based on the intuition that data receiving more positive votes from the labeling sources are, in expectation, more likely to be positive examples.
We formulate this simple idea as constraints on the inferred labels and construct a constrained optimization problem that seeks for a solution satisfying the constraints and minimizing the $\ell$-2 regularization loss.
On 10 benchmark binary classification datasets, we empirically demonstrate the efficacy of \method by showing that it offers highest performance on average.
Specifically, we conduct experiments regarding both the quality of generated labels and the performance of a final classifier supervised with the labels.
In both batches of experiments, we compare \method with modified versions of several existing WS methods as their original implementations typically require LFs for both positive and negative classes.

\section{Related Work}

Weak Supervision is a recent paradigm for building machine learning models without manually labeled training data~\cite{Ratner16, zhang2021wrench}.
It resorts to multiple noisy labeling sources, \eg, heuristics and knowledge bases, to produce probabilistic training labels without hand-labeling.
Specifically, Weak Supervision paradigm abstracts these weak labeling sources as \emph{labeling functions}~\cite{Ratner16}, which are user-defined programs that each provides labels for some subset of the data, collectively generating a large but potentially overlapping and noisy set of votes on training labels.
To aggregate the noisy votes to produce training labels, researchers have developed various \emph{label models}~\cite{Ratner16, Ratner19, fu2020fast, Varma2019multi,zhang2021creating}, which often build on prior work in crowdsourcing, e.g.~\cite{DawidSkene}.
Then, these training labels (often confidence-weighted or probabilistic) are in turn used to train an \emph{end model} which can generalize beyond the labels for downstream tasks.
Recently, Weak Supervision has been widely applied to various machine learning tasks in a diversity of domains: scene graph prediction~\cite{chen2019scene}, video analysis~\cite{fu2019rekall, Varma2019multi}, image classification~\cite{das2020goggles}, image segmentation~\cite{hooper2020cut}, autonomous driving~\cite{Weng2019UtilizingWS}, relation extraction~\cite{Jia2021HeterogeneousGN,zhou2020nero,liu2017heterogeneous}, named entity recognition~\cite{safranchik2020weakly,lison2020named,li2021bertifying, lan2020connet, DBLP:conf/naacl/GoelORVR21}, text classification~\cite{ren2020denoising, yu-etal-2021-fine,shu2020learning,shu2020leveraging},  biomedical~\cite{Kuleshov2019AMD,fries2017swellshark,Mallory2020ExtractingCR}, healthcare~\cite{Fries2021OntologydrivenWS,DBLP:journals/patterns/DunnmonRSKMSGLL20,Fries2019WeaklySC,DBLP:conf/miccai/SaabDGRSRR19,Wang2019ACT,Saab2020WeakSA}. The interested reader is referred to this survey~\cite{zhang2022survey} for more literature review of Weak Supervision.

Existing Weak Supervision studies and applications all assume that labeling sources cover all interested classes, while in some scenarios of binary classification, negative data could be too diverse for users to collect effective labeling sources, \eg, in bot detection the behavioral patterns of normal users could be arbitrary, which also motivates the study of Positive-Unlabeled learning~\cite{upu}.
Therefore, we consider the task of building binary classifier solely based on labeling sources for positive class.
To the best of our knowledge, we are the first that formulates and tackles the problem of learning from only positive labeling sources for binary classification.

\section{Background}

We denote scalar and generic items as lowercase letters, vectors as lowercase bold letters, and matrices as bold uppercase letters. 
% The $i$-th column of a matrix $\mathbf{A}$ is denoted by the corresponding lowercase symbol $\mathbf{a}_i$, \ie, $\mathbf{A}= [\mathbf{a}_1, \mathbf{a}_2, \cdots , \mathbf{a}_n]$.
For a vector $\mathbf{v}$, we use $\mathbf{v}_i$ to represent its $i$-th value.
% Due to space constraints, all proofs are deferred to the appendix.

In binary classification, we aim to learn a scoring function $g$ that could be used to build a binary classifier $h(x)=\text{sign}(g(x)-\pi)$, where $\pi$ is a threshold and $x$ is a data point. In other words, the classifier $h(x)$ maps data $x \in \mathcal{X}$ into binary label $y\in \mathcal{Y}=\{-1, +1\}$.
% A loss function $\ell: \mathbb{R}\times \mathcal{Y} \rightarrow \mathbb{R}_{+}$ quantifies the error of the classifier with respect to the true label $y$.
% Given a classifier $f$, the risk $R(f) \coloneqq \mathbb{E}_{(x, y) \sim p(x, y)} [\ell(f(x), y)]$ quantifies the expected loss of $f$ over $p(x, y)$.
% Since $p(x, y)$ is unknown, \emph{empirical risk} is used in practice.
% Specifically, given a dataset $\mathcal{D} =\{x_i\}_{i\in [N]}$ consisting of $N$ data points and a labeling $\by =[y_1, y_2, ..., y_N]$, the empirical risk is $R(f; \mathcal{D}, \by )=\frac{1}{N}\sum_{i}^{N} \ell(f(x_i), y_i)$.
In standard supervised learning, we are given the ground truth label $\mathbf{y} =[y_1, y_2, ..., y_N]$ of the dataset $\mathcal{D}=\{x_i\}_{i\in [N]}$ for learning an optimal scorer $g$, where $N$ is the size of the dataset.
However, for many classification tasks of interest, the collection of ground truth labels could be expensive and time-consuming.
To tackle the challenge of low availability of ground truth labels, researchers have resorted to Weak Supervision~\cite{Ratner16,zhang2022survey}, which leverages programmatically generated, potentially noisy and correlated labeling sources to synthesize training labels.
In this work, we follow this Weak Supervision setup and do not assume any ground truth label.

% The amount of labeled data required to guarantee that we
% can find (or train) such a classifier is referred to as the sample complexity, which is related to the size or expressivity of
% H. For many classification tasks of interest, there could be
% low availability of labeled data, and this is a critical problem
% for a wide range of domains, where the most successful
% hypothesis classes are very expressive (e.g., convolutional
% neural networks for images).
Formally, we have access to $M$ labeling sources $S=\{\lambda_j\}_{j\in [M]}$.
For concreteness, we follow the general convention of Weak Supervision~\cite{Ratner16} and refer to these sources as \emph{labeling functions (LFs)}.
Different from existing studies of Weak Supervision for binary classification that typically assume LFs could assign positive (+1), negative (-1), or abstain (0) to each data point, we are interested in the setting wherein we do not have negative LFs.
We argue that such a setting is of importance because in real-world scenarios (1) high negative-class diversity may make constructing LFs prohibitively difficult~\cite{DBLP:journals/corr/abs-1811-04820}, or (2) negative data may not be systematically recorded in some domains~\cite{bekker2018estimating} and therefore it is difficult for developers to summarize labeling heuristics.
Thus, in our setting, each LF $\lambda_j$ either assigns positive label ($+1$) to a data or abstains ($0$), resulting a \emph{label matrix} $\mathbf{L}\in \{0, 1\}^{M\times N}$.
Additionally, we assume the class prior $p_{+}=p(y=1)$ is known following the convention of Weak Supervision~\cite{Ratner19}.
We use  $\Lambda(x)$ to represent the output of LFs on data $x$.
We also use $\mathcal{B}^{d}$ to represent $\{0, 1\}^{d}$ and $\Lambda(x)\in \mathcal{B}^{M}$.

Given these LFs, our goal is to learn a \emph{label model} $f_{\theta}(\Lambda(x))$ (short for $f_{\theta}(x)$), which is also a scoring function similar to $g$ but inputs $\Lambda(x)$ instead.
It could be used to either directly make predictions on test data or provide supervisions for training a more complex \emph{end model} $g$ which inputs the data feature, \emph{all without ground truth label}.

\section{The Proposed \method Approach}

\subsection{Conditional Moment Statistics}

First, given a parameterized scoring function $f_{\theta}(x)$ and a possible output of all the LFs $v\in \mathcal{B}^{M}$, we define a moment statistic conditional on $\Lambda(x)=v$ as
\begin{align}
     E_{x \sim \Lambda(x)=v} [ f_{\theta}(x) ],
\end{align}
which is the \emph{averaged score} of $f_{\theta}(x)$ over the set of data where $\Lambda(x)=v$.
Empirically, given the dataset $\mathcal{D}$ at-hand, the conditional moment statistics can be similarly defined as 
\begin{align}
     E_{x_i\in\mathcal{D}, \Lambda(x_i)=v} [ f_{\theta}(x_i) ] =
     E_{x_i\in\mathcal{D}_{v}} [ f_{\theta}(x_i) ] =\frac{1}{|\mathcal{D}_{v}|} \sum_{x_i\in\mathcal{D}_{v}} f_{\theta}(x_i),
\end{align}
where $\mathcal{D}_{v}=\{x_i\in\mathcal{D}|\Lambda(x_i)=v\}$.

\subsection{A Partial Ordering of LFs Output}
Then we introduce the covering relation between two binary vector $v_1, v_2 \in \mathcal{B}^M$.

\begin{definition}[Covering Relation]
\label{def:validgraph}
For $v_1, v_2 \in \mathcal{B}^M$, $v_1$ is covered by $v_2$ if $\forall i \in [M], v_2[i]\geq v_1[i]$ and $\exists j \in [M], v_2[j]> v_1[j]$.
We represent this covering relation using operator $>_{c}$, \eg, $v_2 >_{c} v_1$.
\end{definition}

For example, $v_1$ is covered by $v_2$ if $v_1=[1, 0, 0]$ and $v_2=[1, 1 ,0]$; however for $v_3=[0,0,1]$, there is no covering relation between $v_2$ and $v_3$.
Hence, the covering relation defines a \emph{partial ordering} of elements in $\mathcal{B}^{M}$.

\subsection{Constrained Optimization}

Now we describe our intuition.
We expected that (in expectation) data have more positive votes should be more likely to be positive.
Notably, we do not simply count the number of LFs assigning positive label as LF could be noisy, instead we resort to the covering relation.
The reason is that, consider two possible output of LFs $v_1=[1, 0, 0]$ and $v_2=[0, 1 ,1]$, though $v_2$ has more positive labels, it's not always true to assume $E_{x_i\in\mathcal{D}_{v_2}} [ f_{\theta}(x_i) ]>E_{x_i\in\mathcal{D}_{v_1}} [ f_{\theta}(x_i) ]$ as LFs could have varying noisy level; 
while given $v_1=[1, 0, 0]$ and $v_2=[1, 1, 0]$, it is relatively safer to expect $E_{x_i\in\mathcal{D}_{v_2}} [ f_{\theta}(x_i) ]>E_{x_i\in\mathcal{D}_{v_1}} [ f_{\theta}(x_i) ]$, since the LF assigns positive label to $v_1$ also fires on $v_2$ and $v_2$ has additional supporting LF.
Formally, we expect that 
\begin{align}
     \forall v_1, v_2 \in \mathcal{B}^{M}, v_1 >_{c} v_2, \quad E_{x_i\in\mathcal{D}_{v_1}} [ f_{\theta}(x_i) ] \geq E_{x_i\in\mathcal{D}_{v_2}} [ f_{\theta}(x_i) ]
\end{align}
Note that we only consider $v \in \mathcal{B}^{M}$ if there is at least one $x\in\mathcal{D}$ such that $\Lambda(x)=v$.
Obviously, we could re-write all such inequality in matrix form:
\begin{align}
\label{ineq-mat}
      \mathbf{A} f(\mathbf{x}; \theta)^{\top} \leq \mathbf{0},
\end{align}
where $\mathbf{A}\in\mathbb{R}^{d\times N}$, $d$ corresponds to the number of such inequalities, and $f(\mathbf{x}; \theta)=[f(x_1; \theta), f(x_2; \theta), \cdots , f(x_n; \theta)]$.
Finally, we state our constrained optimization problem which leverages the aforementioned inequality as constraint and $\ell$-2 regularization to learn the label model:
\begin{align}
\label{max-margin}
        \min_{\theta} ||\theta||_{2}^{2} \quad s.t. \quad
      \mathbf{A} f(\mathbf{x}; \theta)^{\top} \leq \mathbf{0},
\end{align}

By absorbing the constraints, we convert the optimization problem into:
\begin{align}
        \min_{\theta} \lambda ||\theta||_{2}^{2} + \sum_{i=1}^{d}\max(
      \mathbf{A}_{i} f(\mathbf{x}; \theta)^{\top}, 0),
\end{align}
where $\lambda$ is the balancing hyper-parameter, which is set to 1 throughout the paper.

\subsection{Parameterization}

We simply parametrize the scoring function $f_{\theta}(x)$ as 
\begin{align}
      f_{\theta}(x) = \Lambda(x) \mathbf{\theta}^{\top},
\end{align}
where $\mathbf{\theta}\in\Delta^{M}$ and $\Lambda(x) \mathbf{\theta}^{\top}$ is a convex combination of LFs output.
Such a simple parametrization restricts the range of $f_{\theta}(x)$ to be $[0, 1]$ and therefore $f_{\theta}(x)$ could be interpreted as $P(y=1|\Lambda(x))$.
Then, we could further incorporate the label prior $p_{+}$ as a constraint.
Specifically, we expect $\frac{1}{N}\sum_{i=1}^{N}f_{\theta}(x_i)=p_{+}$ and the final optimization problem becomes:
\begin{align}
\label{final_obj}
        \min_{\mathbf{\theta}\in\Delta^{M}} \lambda ||\theta||_{2}^{2} + \sum_{i=1}^{d}\max(
      \mathbf{A}_{i} f(\mathbf{x}; \theta)^{\top}, 0) + |\frac{1}{N}\sum_{i=1}^{N}f_{\theta}(x_i)-p_{+}|
\end{align}
% Notably, if we drop the terms corresponding to our constraints, \ie, the second and third term in Eq.~\ref{final_obj}, the optimization problem will naturally lead to solution $\theta=[\frac{1}{M}, \frac{1}{M}, \cdots, \frac{1}{M}]$, which is simply the Majority Voting method.
Such an optimization problem can be readily and efficiently solved by existing library, \eg, CVXPY~\cite{diamond2016cvxpy}.

\section{Experiments}

\subsection{Datasets}

Throughout the experiments, we use the following 10 binary classification datasets from WRENCH~\cite{zhang2021wrench}, a comprehensive benchmark platform for Weak Supervision: \textbf{Census}, \textbf{Mushroom}, \textbf{Spambase}, \textbf{PhishingWebsites}, \textbf{Bioresponse}, \textbf{BankMarketing}, \textbf{CDR}, \textbf{SMS},  \textbf{Yelp},  and \textbf{IMDb}.
Note that the first 6 datasets are tabular dataset while the remaining ones are textual datasets.
For all the datasets, we only use the positive labeling functions provided by WRENCH.
For textual datasets, we use pretrained BERT model~\cite{devlin2019bert} to extract features.

\subsection{Compared Methods}

We compare \method, as well as its variant that does not leverage the label prior (\method-prior), with the following label models in the literature.
\textbf{MV:} We adopt the classic majority voting (MV) algorithm as one label model. Notably, the abstaining LF, \ie, $\lambda_j = 0$ won't contribute to the final votes.
\textbf{DS~\cite{DawidSkene}:} Dawid-Skene (DS) model estimates the accuracy of each LF with expectation maximization (EM) algorithm by assuming a naive Bayes distribution over the LFs’ votes and the latent ground truth.
\textbf{MeTaL~\cite{Ratner19}:} MeTal models the distribution via a Markov Network and recover the parameters via a matrix completion-style approach.
\textbf{FS~\cite{fu2020fast}:} FlyingSquid (FS) models the distribution as a binary Ising model, where each LF is represented by two random variables.
A Triplet Method is used to recover the parameters and therefore no learning is needed, which makes it much faster than DS and MeTal.
However, all existing label models assume the presence of negative LF that is absent in our setting.
Thus, we treat the abstain (0) as negative (-1) so that existing label models are applicable.

\subsection{Evaluation Protocol}

First, we compare the performance of label models on test set. Notably, there is a subset of data not covered by any LF and therefore the label models have no information on them.
Thus, we only evaluate the performance of label models on the covered subset of test data.
We also found it is beneficial to treat these uncovered data as negative example when training the end model, since covered data are more likely to be positive, leaving the uncovered more likely to be negative.

Then, we compare the performance of end model trained with signals produced by label model. 
Notably, throughout the experiments, we do not use any clean labels as validation set for model selection,  as it contradicts to our setting of absence of clean labels.
We found that in such a setup the kernel ridge regression model with RBF kernel \footnote{\url{https://scikit-learn.org/stable/modules/generated/sklearn.kernel_ridge.KernelRidge}} outperforms other options, \eg, linear regression, logistic regression, multi-layer perceptron classifier and so on.

For both evaluations, we adopt two common metrics for binary classification, namely, the Area Under the Receiver Operating Characteristic Curve (ROC-AUC score) and the Area Under the Precision-Recall curve (PR-AUC score), since they can be used to directly evaluate the scoring function $f_{\theta}(x)$.

\subsection{Results: Label Models}

The main results of label model comparison are presented in Table~\ref{tab:labelmodel}.
First of all, we can see that \method achieves the highest averaged performance under both evaluation metrics and its variant \method-prior has the second best performance.
This demonstrates the efficacy of \method as well as incorporating label prior as a constraint.
However, \method does not outperform all the baselines on every datasets, which aligns with the recent results in a benchmarking study~\cite{zhang2021wrench} that it is unlikely to have a universally-best label model.

\begin{table*}[t]
	\centering
	\caption{Label model comparison on covered test data. We highlight the best performance across Weak Supervision method in bold.}
	\scalebox{0.7}{ 
        \begin{tabular}{c|c|cccccccccccc}
        		\toprule
        		 \multicolumn{2}{c|}{\multirow{1}{*}{\textbf{Method}}} & 
        		 \textbf{Census}  & \textbf{Mushroom}  & \textbf{Spambase}  & \textbf{PhishingWebsites}  & \textbf{Bioresponse}  & \textbf{BankMarketing}  & \textbf{CDR} & \textbf{SMS} &  \textbf{Yelp}  & \textbf{IMDb}  &  \textbf{Average} \\
    \midrule
	
	\multicolumn{13}{c}{\textbf{ROC-AUC score}}\\\midrule
	
	\multicolumn{2}{c|}{MV} & 58.28 & 66.02 & 68.25 & 65.41 & 57.73 & 67.97 & 58.81 & 40.00 & 66.31 & 55.59 & 60.44 \\

	\multicolumn{2}{c|}{DS} & 37.19 & 66.71 & 64.73 & 30.96 & 57.99 & 48.92 & 47.26 & 37.50 & \textbf{69.13} & \textbf{61.86} & 52.22\\
	
	\multicolumn{2}{c|}{Snorkel} & 52.07 & 53.05 & 68.81 & 38.47 & 57.59 & 63.44 & 57.54 & 32.50 & 68.90 & 50.06 & 54.24 \\

	\multicolumn{2}{c|}{FS} & 56.76 & 65.91 & \textbf{75.27} & 53.97 & \textbf{63.71} & \textbf{71.12} & 55.98 & 32.50 & 64.14 & 49.05 & 58.84 \\
	
	\midrule
	
	\multicolumn{2}{c|}{\method-prior} & \textbf{64.91} & 62.58 & 71.10 & 57.52 & 59.21 & 62.98 & 62.01 & 72.50 & 67.80 & 50.07 & 63.07 \\
	
	\multicolumn{2}{c|}{\method} & 56.72 & \textbf{68.58} & 75.15 & \textbf{67.22} & \textbf{63.71} & 64.88 & \textbf{62.94} & \textbf{75.83} & 65.64 & 61.19 & \textbf{66.19}\\
	
	\midrule\multicolumn{13}{c}{\textbf{PR-AUC score}}\\\midrule
    
    \multicolumn{2}{c|}{MV} & 66.59 & 82.67 & 86.82 & 75.75 & 71.39 & 32.92 & 59.18 & 19.36 & 75.48 & 71.57 & 64.17 \\

	\multicolumn{2}{c|}{DS} & 54.32 & 85.22 & 87.13 & 58.03 & 73.00 & 26.31 & 53.23 & 20.29 & \textbf{80.93} & \textbf{75.47} & 61.39\\
	
	\multicolumn{2}{c|}{Snorkel} & 67.24 & 83.76 & 88.26 & 67.92 & 72.52 & 30.81 & 60.17 & 19.08 & 80.21 & 72.44 & 64.24\\

	\multicolumn{2}{c|}{FS} & 70.42 & 86.86 & \textbf{91.38} & 71.93 & \textbf{76.18} & \textbf{36.43} & 59.74 & 19.08 & 76.31 & 70.76 & 65.91  \\
	
	\midrule
	
	\multicolumn{2}{c|}{\method-prior} & \textbf{71.61} & 85.74 & 90.88 & 74.93 & 74.96 & 30.78 & 61.44 & 49.38 & 79.21 & 72.43 & 69.13 \\
	
	\multicolumn{2}{c|}{\method} & 70.24 & \textbf{88.14} & 91.33 & \textbf{78.32} & \textbf{76.18} & 31.64 & \textbf{61.85} & \textbf{66.52} & 76.85 & 74.32 & \textbf{71.54} \\
    
		\bottomrule
         \end{tabular}
    }
    \label{tab:labelmodel}
\end{table*}

\subsection{Results: End Models}

\begin{table*}[h]
	\centering
	\caption{End model comparison on test data. We highlight the best performance across Weak Supervision method in bold.}
	\scalebox{0.7}{
        \begin{tabular}{c|c|cccccccccccc}
        		\toprule
        		 \multicolumn{2}{c|}{\multirow{1}{*}{\textbf{Method}}} & 
        		 \textbf{Census}  & \textbf{Mushroom}  & \textbf{Spambase}  & \textbf{PhishingWebsites}  & \textbf{Bioresponse}  & \textbf{BankMarketing}  & \textbf{CDR} & \textbf{SMS} &  \textbf{Yelp}  & \textbf{IMDb}  &  \textbf{Average} \\
    \midrule
	
	\multicolumn{13}{c}{\textbf{ROC-AUC score}}\\\midrule
	
	\multicolumn{2}{c|}{Gold} & 89.03 & 100.00 & 96.56 & 99.24 & 80.70 & 91.10 & 81.21 & 99.90 & 95.67 & 89.32 & 92.27 \\
	
	\multicolumn{2}{c|}{Gold-covered} & 83.71 & 100.00 & 89.40 & 97.90 & 72.96 & 89.88 & 81.13 & 98.40 & 94.81 & 88.75 & 89.69 \\
	
	\multicolumn{2}{c|}{Gold--slice} & 81.96 & 98.32 & 85.26 & 90.10 & 70.11 & 62.63 & 78.12 & 98.30 & 88.33 & 85.51 & 83.86 \\
	
	\multicolumn{2}{c|}{Gold--train} & 72.75 & 87.63 & 83.54 & 84.14 & 69.76 & 62.18 & 74.91 & 96.16 & 90.07 & 85.42 & 80.66 \\\midrule
	
% 	\multicolumn{2}{c|}{POnce} & 80.49 & \textbf{92.38} & 82.73 & 74.80 & 69.50 & 80.04 & 77.64 & \textbf{98.30} & 85.33 & 84.40 & 82.56\\

	\multicolumn{2}{c|}{MV} & 78.74 & 90.85 & 82.66 & \textbf{78.08} & 69.71 & \textbf{84.48} & 76.95 & 92.92 & 85.68 & 84.38 & 82.45 \\

	\multicolumn{2}{c|}{DS} & 76.55 & 91.02 & \textbf{83.33} & 60.21 & 69.33 & 76.89 & 75.27 & 90.76 & \textbf{87.68} & \textbf{85.35} & 79.64 \\
	
	\multicolumn{2}{c|}{Snorkel} & \textbf{80.50} & 85.25 & 82.95 & 67.62 & 69.62 & 80.66 & 77.24 & \textbf{97.80} & 85.79 & 83.35 & 81.08 \\

	\multicolumn{2}{c|}{FS} & 80.48 & \textbf{92.21} & 82.89 & 72.27 & 69.71 & 82.06 & 77.29 & 97.63 & 85.34 & 83.93 & 82.38 \\
	
	\midrule
	\multicolumn{2}{c|}{\method-prior} & 78.30 & 90.59 & 82.44 & 77.37 & \textbf{69.79} & 83.87 & 73.96 & 96.03 & 85.59 & 83.88 & 82.18 \\

	\multicolumn{2}{c|}{\method} & 79.12 & 91.98 & 83.15 & 78.01 & 69.50 & \textbf{84.48} & \textbf{78.01} & 96.82 & 86.40 & 84.93 & \textbf{83.24} \\
	
	\midrule\multicolumn{13}{c}{\textbf{PR-AUC score}}\\\midrule
    
     \multicolumn{2}{c|}{Gold} & 73.90 & 100.00 & 96.61 & 99.44 & 79.96 & 55.40 & 65.86 & 99.39 & 96.16 & 89.05 & 85.58  \\
     
     \multicolumn{2}{c|}{Gold-covered} & 68.78 & 100.00 & 88.24 & 98.50 & 74.77 & 53.11 & 66.30 & 94.21 & 95.48 & 88.73 & 82.81 \\
     
     \multicolumn{2}{c|}{Gold--slice} & 65.44 & 96.06 & 83.54 & 91.72 & 70.54 & 33.78 & 60.19 & 93.76 & 89.55 & 85.54 & 77.01\\
     
     \multicolumn{2}{c|}{Gold--train} & 54.28 & 82.33 & 81.09 & 86.70 & 71.40 & 31.71 & 58.70 & 88.48 & 91.25 & 85.44 & 73.14\\\midrule
	
% 	\multicolumn{2}{c|}{POnce} & 57.81 & 83.73 & 77.90 & 72.74 & 70.06 & 25.41 & 59.46 & \textbf{93.76} & 86.79 & 84.08 & 71.17\\

	\multicolumn{2}{c|}{MV} & 57.10 & 85.06 & 80.12 & 79.23 & 71.92 & \textbf{36.38} & 59.00 & 84.00 & 87.04 & 84.06 & 72.39 \\

	\multicolumn{2}{c|}{DS} & 47.22 & 85.70 & \textbf{81.54} & 62.37 & 70.37 & 26.78 & 53.86 & 81.47 & \textbf{89.12} & \textbf{85.25} & 68.37 \\
	
	\multicolumn{2}{c|}{Snorkel} & \textbf{57.92} & 74.65 & 80.21 & 71.07 & 71.57 & 28.96 & 58.95 & \textbf{92.70} & 87.24 & 82.79 & 70.61 \\

	\multicolumn{2}{c|}{FS} & 57.89 & 88.50 & 79.23 & 71.01 & 71.56 & 33.68 & 59.00 & 91.81 & 86.78 & 83.52 & 72.30 \\
	
	\midrule
	
	\multicolumn{2}{c|}{\method-prior} & 56.80 & 85.03 & 80.64 & \textbf{79.55} & \textbf{72.01} & 35.46 & 55.25 & 86.29 & 86.44 & 83.49 & 72.10 \\
	
	\multicolumn{2}{c|}{\method} & 57.20 & \textbf{88.99} & 80.35 & 78.89 & 70.94 & 36.37 & \textbf{60.74} & 89.30 & 87.89 & 84.72 & \textbf{73.54}\\
    
		\bottomrule
         \end{tabular}
    }
    \label{tab:endmodel}
\end{table*}

For evaluation of end models, we additionally include four methods involving ground truth label for understanding the upper-bound performance we could achieve.
\textbf{Gold}: it use all the ground truth labels of training data to train the end model;
\textbf{Gold-cover}: it use the ground truth labels of \emph{covered} training data to train the end model, since for uncovered data we do not have information for their underlying labels as no LF fires;
\textbf{Gold-slice}: because label model inputs only the votes of LF, a group of data sharing the same votes would receive the same output from the label model, Gold-slice assigns the most-likely label for such a group of data, which is more close to the upper-bound performance any label model could offer;
\textbf{Gold-train}: it use all the ground truth label to train a label model $f_\theta$ with the same model class as \method, \ie, $\mathbf{\theta}\in\Delta^{M}$.

For the results in Table~\ref{tab:endmodel}, we can first see that \method still achieve the highest averaged performance and again, it is not the best method for every dataset as observed in the label model comparison.
Surprisingly, we found that \method is slightly better than Gold-train in terms of averaged performance, which further indicates the efficacy of our design.
Finally, we conclude that even without negative LFs, Weak Supervision paradigms could still render reasonably good performance for binary classification, as the gap between Gold method and compared Weak Supervision methods is around 10\%.

\section{Conclusion}
In this work, we study Weak Supervision for binary classification with positive labeling sources only.
We propose \method, a simple yet effective method for such a novel setting.
It leverages the intuition that data receiving more votes from positive labeling sources are in expectation more likely to be positive.
Empirically, we compare \method with several baselines modified for this setting and show that it offers highest averaged performance.
\bibliographystyle{ACM-Reference-Format}
\bibliography{cikm2022}

\end{document}